\documentclass{IOS-Book-Article}

\usepackage{mathptmx}
\usepackage{array}
\usepackage{subcaption}

\newcolumntype{x}[1]{>{\centering\arraybackslash\hspace{0pt}}p{#1}}

\def\hb{\hbox to 10.7 cm{}}

\begin{document}

\begin{frontmatter}              

\title{Advanced Rich Transcription System for Estonian Speech}
\runningtitle{Advanced Rich Transcription System for Estonian Speech}


\author{\fnms{Tanel} \snm{Alum\"{a}e}%
\thanks{Corresponding Author: Tanel Alum\"{a}e; E-mail:
tanel.alumae@ttu.ee. \\
This research has been supported by the Centre of Excellence in Estonian
Studies (CEES, European Regional Development Fund).}
},
\author{\fnms{Ottokar} \snm{Tilk}}
and
\author{\snm{Asadullah}}

\runningauthor{T. Alum\"{a}e  et al.}
\address{Laboratory of Language Technology, Tallinn University of Technology, Estonia}

\begin{abstract}
This paper describes the current TT\"{U} speech transcription system for Estonian speech. The system is designed to handle  semi-spontaneous speech, such as broadcast conversations, lecture recordings and interviews recorded in diverse acoustic conditions. The system is based on the Kaldi toolkit. Multi-condition training using background noise profiles extracted automatically from untranscribed  data is used to improve the robustness of the system. Out-of-vocabulary words are recovered using a  phoneme $n$-gram based decoding subgraph  and a FST-based phoneme-to-grapheme model. 
The system achieves a word error rate of 8.1\% on a test set of broadcast conversations.
The system also performs punctuation recovery and speaker identification. 
Speaker identification models are trained using a recently proposed weakly supervised training method.
\end{abstract}

\end{frontmatter}

\section{Introduction}

An automatic speech recognition (ASR) system converts a speech recording to a stream of orthographic words. Various technologies can be applied to enrich such word streams, such as attributing speech segments to different speakers, identifying the speakers by name, dividing the word stream into sentences and adding punctuation symbols. Rich transcription makes ASR results more readable and valuable for human users. It also enhances the content for various down-stream natural language processing (NLP) applications, such as spoken document retrieval, summarization, machine translation, semantic navigation, speech data mining, and others.

This paper describes the recent improvements to Tallinn University of Technology (TT\"{U}) Estonian speech transcription system. The system is designed to handle mainly semi-spontaneous speech from various domains, such a broadcast conversations, lecture recordings and interviews. Previous versions of the system have been described in \cite{alumae2014recent} and \cite{alumae2012transcription}. One current focus of the system is better handling of data recorded ``in the wild'', such as interviews and meetings recorded in adverse real world acoustic conditions. In addition to speech-to-text, the system also performs automatic punctuation restoration and speaker identification. The speaker identification system can identify a wide range of public figures by voice and is trained in a weakly supervised manner.

The described system is free and open source \footnote{http://github.com/alumae/kaldi-offline-transcriber}. It is used as the backend for our public web-based speech transcription service \footnote{http://bark.phon.ioc.ee/webtrans} and by several Estonian media-monitoring companies for transcribing radio and TV broadcasts.

\section{Training data}

\subsection{Speech data}

Speech data that is used for training the acoustic models is summarized in Table \ref{tbl:speech_data}. Only the duration of the  segments containing transcribed speech is shown, i.e., segments containing music, long periods of silence or are left untranscribed is excluded.

\begin{table}[tb]
    \caption{ Training data for speech recognition.}
\begin{subtable}{.49\linewidth}
\caption{\label{tbl:speech_data}Acoustic model training data.}
\begin{tabular}{lr}
\hline
Source  & Amount (h) \\
\hline
Broadcast conversations &  114.0 \\
Broadcast news &  37.2 \\
Conference speeches, lectures &  37.9 \\
Spontaneous speech \cite{lippus2011acoustic} &  39.6 \\
Parliament speeches  & 31.0 \\
BABEL speech database \cite{eek1999estonian}  & 7.0 \\
Other &  1.8 \\
\hline
Total &  268.5 \\
\hline
\end{tabular}
\end{subtable}
\begin{subtable}{.49\linewidth}
\caption{\label{tbl:text_data} Language model training data.}
\begin{tabular}{lr}
\hline
Source & Tokens (M) \\
\hline
Web & 434 \\
Newspapers & 196 \\
Magazines and journals & 28 \\
Parliament transcripts & 15 \\
Social media (blogs, comments) & 9.4 \\
Broadcast conversation transcripts & 0.98 \\
Broadcast news transcripts & 0.33 \\
Lecture and conference transcripts & 0.28 \\
\hline
Total &  690 \\
\hline
\end{tabular}
\end{subtable}
\end{table}

Most of the speech data (broadcast speech, conference and lecture speeches, parliament speeches) used for training are collected in Tallinn University of Technology during the past 10 years \cite{meister2012new}.

In addition to transcribed speech data, we make use of large amounts of untranscribed speech data. The data originates from real usage of our public web-based speech transcription service. The service is used mostly for transcribing lectures, interviews, meetings and other types of mostly semi-spontaneous speech. The user-made audio recordings are often recorded in noisy environments, using a microphone positioned relatively far from the user (e.g., a smart phone lying on the table). Therefore, this kind of data has often significant reverberation and background noise. The data was randomly selected from the 2017 service usage data. To eliminate very random usage data, the recordings were picked from the recordings uploaded by about 50 most active service users. The total number of recordings is 2122 and the total duration is 908 hours. We use this data for automatically extracting background noise segments, which in turn are used as in-domain training data for noise augmentation (see Section \ref{sec:am}).

\subsection{Text data}

Text data sources used for training the language models
(LMs) are listed in Table \ref{tbl:text_data}, with the number of tokens in each source after normalization and compound word splitting. Most of the written language
corpora are compiled at the University of Tartu \cite{kaalep2005corpora}. In order
to have up-to-date language data, we scrape additional
web data from news portals and blogs. Finally, transcriptions
of conversational broadcast data (talk shows, telephone interviews) and conference speeches are used as a sample of spoken language.

Before using the text data for LM training, text normalization
is performed. Texts are tokenized, split into sentences
and recapitalized, i.e., converted to a form where names and
abbreviations are correctly capitalized while normal words at
the beginning of sentences are written in lower case. 
Recapitalization is performed using a simple count-based model that converts a capitalized word at the beginning of the sentence to its most common intra-sentence form. For expanding
numbers into words, a non-trivial approach is needed as
the exact textual representation of each number depends on
the inflection. However, the inflection of a number is usually
not visible in orthography and is inferred from the context
by human readers. To determine the inflection of a written
number, we employ a semi-supervised machine learning approach
similar to [8]: a support vector machine classifier is
first built using the numbers that are already written as words
in training texts, and the classifier is then used to determine
the inflection for the rest of the numbers. Neighboring words
and their suffixes are used as features for the classifier. 

As Estonian is a heavily compounding and inflective language,
the lexical variety of the language is very high. To reduce
the out-of-vocabulary (OOV) rate of the LM, compound
words are decomposed into compound segments, using the
word structure information assigned by a morphological analyzer
\cite{kaalep2001complete}. Compound words are later reconstructed from the output of the speech recognition system using a hidden-event n-gram model \cite{alumae2007automatic}.

\section{Speech recognition system}

Our speech recognition system is based on the Kaldi toolkit \cite{kaldi}. It incorporates many novel techniques recently implemented in Kaldi, such as factored time-delay neural network (TDNN-F) acoustic models and a neural network language model that uses words and character-based features. In the following, we focus on the details that are different from the standard Kaldi recipes.

The system also includes a speaker diarization module based on the LIUM SpkDiarization toolkit \cite{meignier2010lium}  and a rule-based pronunciation dictionary. They haven't significantly changed since 2014 and are described in detail in \cite{alumae2014recent}.

\subsection{Acoustic modeling}
\label{sec:am}

Our neural network acoustic model uses the recently introduced TDNN-F architecture \cite{povey2018semi} and is trained with lattice-free MMI criterion \cite{povey2016purely}. The hyper-parameters of the training setup are mostly borrowed from the best-performing Kaldi Switchboard recipe, as it uses similar amount of training data. I-vector based speaker adaptation is used.

To make the system more robust towards adverse acoustic conditions, we use heavy data augmentation for training the neural network acoustic model. We use nine-fold data augmentation: the original training data is three-fold speed-perturbed (using speedup factors 0.9, 1.0 and 1.1) and volume-perturbed. A second copy of the training data is artificially reverberated with various small and medium room impulse responses and mixed with various environmental background noises from the MUSAN corpus \cite{musan}. Similarly to the clean data, this noise-augmented copy is further three-fold replicated using noise and volume perturbation. This is based on the approach implemented in the Kaldi recipes for multi-condition training \cite{ko2017study}. 
rs
To further adapt the acoustic models for real-world acoustic conditions, we create a second reverberated and noisy copy of the clean training data. Instead of using various background noises from an external corpus, we extract the non-speech sections from our untranscribed real usage data, using a speed activity detection (SAD) system. We first train a deep-neural network based SAD model on noise-augmented transcribed clean speech data, and then use it to extract non-speech segments from the untranscribed noisy speech data. 
Non-speech segments extracted from individual recordings are concatenated and used as background noises for the second noise augmentation round. Details of this method can be found in \cite{asadullah2018}. 
As a result, the acoustic model is trained on nine copies of the original training data (clean, noise-augmented with external noises, noise-augmented with in-domain noises, each with three-fold speed and volume perturbation). 

\subsection{Language modeling}

Language model (LM) vocabulary is created by selecting the 200\,000 most
likely case-sensitive compound-split units from the unigram
mixture of the individual training corpora, optimized on the development data.
For each corpus, a 4-gram LM is built, using interpolated modified Kneser-Ney
discounting. The individual LMs are interpolated into
one by using interpolation weights optimized on development
data. Finally, the LM is heavily pruned to less than one tenth
in size using entropy pruning. Even more aggressively pruned LM is created for the
first pass of decoding.

We also use a recurrent neural network LM (RNNLM), trained with the recent Kaldi implementation \cite{xu2018neural}.
Since there is no proper way to statically interpolate neural network LMs, we employ the following method to create optimally balanced training data for the neural network LM, in order to avoid biasing it towards written data when all data would be pooled: we define a maximum number of sentences for training a RNNLM, $N_{rnnlm}$ (with $N_{rnnlm}=3\,000\,000$ in the reported experiments) and then subsample  or oversample sentences from the individual LM training corpora so that the number of the retained sentences from each of the corpora would be proportional to the optimal LM interpolation weights that were computed earlier when creating $N$-gram language models. We also define an upper limit $f_{max}=10$ for oversampling and a smoothing factor $\beta=0.5$ to make the sampling factors less peaky. The final sampling factor for each LM subcorpus is calculated as:
\[
f_{i} = min \left( f_{max}, { \left( \frac{w_i * N_{rnnlm}}{N_i}\right) }^\beta \right)
\]
where $w_i$ is the optimal LM interpolation weight for subcorpus $i$ and $N_i$ the number of total sentences in subcorpus $i$. 

\subsection{Recovering out-of-vocabulary words}

Although we perform compound splitting and use a large vocabulary, speech still contains words that are not covered by our language model vocabulary. This includes both proper names not seen in language model training data, as well as common nouns, verbs and adjectives in rare inflections. In order to improve the recognition of such OOV words, we use a special decoding graph modification recently implemented in Kaldi: the OOV words are represented in the decoding graph using an $n$-gram phoneme language model, estimated on the pronunciations of all in-vocabulary words. After decoding, the most likely phoneme sequence corresponding to recognized OOV words can be reconstructed. To turn the phoneme sequence into a word, we use an approach based on finite state transducers (FSTs). We manually constructed an FST that represents Estonian letter-to-phoneme rules, using the Pynini toolkit \cite{gorman2016pynini}. Most of the rules handle context-sensitive rewrite rules for plosive phonemes, but there are also rules for deriving the Estonian pronunciation for common foreign consonant clusters, so that a name \textit{Chris} is transformed to a pronunciation $/k r i s/$ and so on. The FST framework allows to easily invert a transducer, which can be used to convert the grapheme-to-phoneme converter into a phoneme-to-grapheme converter. However, converting phonemes to graphemes is much more ambigious: for example, the phoneme sequence $/k r i s/$ can correspond (according to the inverted FST) to words \textit{kris}, \textit{krys}, \textit{Grys}, \textit{Chriz}, \textit{criz} and many others, without any ranking. To disambiguate between such variants, we compose the result of the phoneme-to-grapheme translation with another FST that represents a grapheme 5-gram model, estimated over all in-vocabulary words. This assigns a higher probability to those words that contain letter sequences which occur more often in in-vocabulary words. The final recovered word is simply the one with the highest probability. 

The described implementation is available at https://github.com/alumae/et-g2p-fst.

\subsection{Experimental results}

We measure speech recognition word error rate (WER) in three domains: broadcast conversations, consisting mostly of radio talkshows, speeches of a linguistics conference and a set of user recordings recorded ``in the wild''. The first two of them are described in detail in \cite{alumae2014recent}. The third set consists about five hours of randomly selected user data uploaded to our public web-based transcription service.  The development set is about two hours and a test set of about three hours in length. 

\begin{table}[tb]
\caption{\label{tbl:wer} Word error rates for different types of test data, using the full system and with one of the proposed improvements deactivated.}
\centerline{	
\begin{tabular}{l|cc|cc|cc|>{\centering\arraybackslash}p{1.3cm}}
\hline
System &  \multicolumn{2}{>{\centering\arraybackslash}p{1.7cm}|} {Broadcast conversations} & \multicolumn{2}{>{\centering\arraybackslash}p{1.7cm}|} {Conference speeches} &   \multicolumn{2}{>{\centering\arraybackslash}p{1.7cm}|} {User recordings} & Average relative WER incr.
\\
& Dev & Test & Dev & Test & Dev & Test \\
\hline
\textbf{Full system} & \textbf{11.0}	 & \textbf{8.1} & \textbf{14.5} & \textbf{12.9} & 29.4 & \textbf{22.7} & \\
6-fold data augmentation (not 9) & 11.0 & 8.2 & 14.9 & 13.4 & 30.1 & 23.0 & +1.9\% \\
No OOV model & 11.3 & 8.3 & 15.2 & 14.0 & \textbf{ 29.2 }& 23.1 & +3.3\% \\
No RNNLM rescoring & 11.8 & 9.0 & 15.9 & 14.0 & 31.2 & 24.6 & +8.5\% \\
\hline
2014 system \cite{alumae2014recent} & 18.0 & 17.9 & 23.7 & 26.3 & N/A & N/A & +88\%\\
\hline
\end{tabular}}
\end{table}

Table \ref{tbl:wer} lists WER results of the ablation study, showing the performance of the full system and with either of the three system components removed. It also shows the reported WER results in \cite{alumae2014recent} for the first two domains. Average relative deterioration of the WER, compared to the full system, is given in the last column.

As expected, the nine-fold data augmentation with in-domain noise (as opposed to six-fold data augmentation) has the biggest effect in the more difficult domains (conference speeches and user recordings). It doesn't deteriorate the results in broadcast speech where the acoustic conditions are mostly clean.

The proposed OOV recovery approach reduces WER in most cases. Analyzing the actual recognition hypotheses shows that this method can greatly improve the readability of transcripts containing OOV words. Although the phoneme-based OOV word reconstruction is not always accurate (e.g., OOV foreign names are transcribed using Estonian pronunciation rules), it is cognitively easier to understand than the text produced by the earlier system where OOV words are replaced by acoustically and lexically similar in-vocabulary words. Some examples of successful and failed OOV recoveries are given in Table \ref{tbl:oov_recovery}.

\begin{table}[tb]
\caption{\label{tbl:oov_recovery} Examples of words that were recognized using the proposed OOV-recovery system. The reconstructed words are written in bold.}
\centerline{	
\begin{tabular}{p{3.5cm}|p{3.5cm}|p{3.5cm}}
\hline
Reference & Without OOV recovery & With OOV recovery \\
\hline
\textit{valdkondlikult} &\textit{valdkond likult} & \textit{\textbf{valdkondlikult}} \\
\textit{sugu m\"{a}rkivatest} & \textit{sugum\"{a}rki vetes} & \textit{sugu \textbf{m\"{a}rkivatest}} \\
\textit{liibukad retuusid }& \textit{liibutav viiret uusi} & \textit{\textbf{liibuked retoosid}} \\
\textit{hingamiseta pausid} & \textit{hingamispausid} & \textit{\textbf{hingamisetabausid}} \\
\textit{t\"{u}lis ja n\"{a}\"{a}kluses} & \textit{t\"{u}lis ja n\~{o}rkuses }& \textit{\textbf{t\"{a}lise n\"{a}\"{a}kluses}} \\
\hline
\end{tabular}}
\end{table}

\section{Punctuation recovery system}
To improve readability of ASR output we restore commas, periods and question marks using a bidirectional RNN punctuation recovery model~\cite{tilk2016brnn}, with improvements described below.

Firstly, due to inflection and compounding, the used 100000 word vocabulary provides poor
coverage. As a result, many words are mapped to a shared UNK token. To produce more informative
embeddings for the out-of-vocabulary words, we train a character based RNN on the existing
100000 word embeddings to generate new embeddings on the fly. The approach is inspired from
\cite{pinter2017}, but our model is unidirectional as our experiments did not show improvements with a
bidirectional model, and uses gated recurrent units \cite{cho2014gru} instead of LSTM.

Second problem is that the original version of the punctuation restoration model \cite{tilk2016brnn}
segments text into fixed length sequences and punctuation restoration is performed on full sequences.
The problem with this approach is that close to the end of the sequence the number of forward
context words diminishes (zero at the last word) reducing the utility of the backwards recurrence.
Sufficient forward context is important for accurate predictions, as is evident from the significantly 
inferior performance of unidirectional models \cite{tilk2015lstm}. We propose a simple fix --- use the 
full sequence as input to the model, but treat the last $n$ words as padding and do not attempt to 
generate punctuation for them.

\subsection{Experimental results}
The experiments use the pre-trained Estonian models from ~\cite{tilk2016brnn}. The embedding generator has 256 hidden units and an input vocabulary of 73 characters plus one UNK symbol. Word embeddings are grouped by word length into minibatches of up to 64 items and 10\% of minibatches are used for validation. The rest of the training settings are identical to the punctuation recovery model ~\cite{tilk2016brnn}.
For sequence padding we set $n=10$.

On the manually transcribed reference transcripts, the proposed methods improve the F1-score
by 0.5-1.1\% and slot error rate (SER) by 1.3-2.2\% relative. On the ASR output
the relative improvements are 0.5-0.6\% in F1-score and 0.1-0.2\% in SER. Smaller improvements
on the ASR output can partially be explained by the fact that the ASR system produces a smaller
amount of unknown words (4.7\% in ASR output vs. 6.3\% in manually transcribed text), making
the generated embeddings less useful. Both proposed approaches can be applied to already existing
models without retraining.

\section{Speaker identification system}

The transcription system performs speaker identification, covering a a wide set of Estonian public figures, such as politicians, state officials, scientists and artists. 

Conventional speaker identification models are usually trained on data where the speech segments corresponding to the target speakers are hand-annotated. However, the process of hand-labelling speech data is expensive and doesn't scale well, especially if a large set of speakers needs to be covered. Instead, we use a method to train speaker identification models using only the information about speakers appearing in each of the recordings in training data, without any segment level annotation \cite{karu2018}.  Obtaining or creating such training data is much easier than segment-annotated data. We use 6000 recordings (originating from between 2004 to 2016) of the main evening news programme \textit{P\"{a}evakaja} of Estonian Public Broadcasting as training data. Most recordings in the archive are accompanied with metadata that lists all speakers (both news reporters and interviewees) speaking in that programme. We use the weakly-supervised training method to construct models for almost 5000 unique speakers. Evaluation is performed on 16 manually annotated recordings of the same news programme from a period not covered by the training set. 

Evaluation reveals that the method results in a 45\% recall rate of speakers appearing in new news programmes, using a 95\% precision threshold, given reference speaker segmentations.
Table \ref{tbl:sid} lists time-weighted identification error rate (IER), precision and recall values after oracle and automatic diarization. A detailed description of the method and the results are given in \cite{karu2018}.

\begin{table}[tbh]
	\caption{\label{tbl:sid}Time-weighted speaker identification results on manually segmented and automatically segmented \textit{P\"{a}evakaja} evaluation set.}
	\label{tbl:paevakaja_results2}
	\vspace{2mm}
	\centerline{
		\begin{tabular}{|l|c|c|c||c|c|c|}
			\hline
			\textbf{Speakers} & \textbf{IER} & \textbf{Precision} & \textbf{Recall} & \textbf{IER} & \textbf{Precision} & \textbf{Recall} \\
			\hline
			\hline
			& \multicolumn{3}{l||}{\textit{Oracle diarization}} & \multicolumn{3}{l|}{\textit{Automatic diarization}} \\
			\hline
			All & 28\% & 96\% & 75\% & 35\% & 93\% & 66\% \\
			\hline
			Anchors & 4\% & 98\% & 98\% & 14\% & 94\% & 89\% \\
			\hline
			Non-anchors & 78\% &  89\% & 24\% & 80\% &  90\% & 22\%  \\
			\hline
		\end{tabular}}
	\end{table}
    
\section{Conclusion}

This paper described the current state of the TT\"{U} Estonian speech transcription system. The system achieves 8.1\% WER on a test set of broadcast conversations, 12.9\% WER on conference speeches and 22.7\% WER on various user-made recordings ``from the wild''. The system also performs punctuation recovery and speaker identification. The speaker identification models are created using weakly supervised training and achieve 66\% time-weighted speaker recognition recall at 93\% precision on a broadcast news test set.

%

\bibliography{main}
\bibliographystyle{ieeetr}

\end{document}